\documentclass[sn-mathphys]{sn-jnl}

\usepackage{graphicx}
\usepackage{float}
\usepackage{natbib}

\usepackage{dirtytalk}
\usepackage{amsmath}

\jyear{2021}%

\theoremstyle{thmstyleone}%
\theoremstyle{thmstyletwo}%

\theoremstyle{thmstylethree}%

\DeclareMathOperator{\ata}{ATA(S)}
\DeclareMathOperator{\segmentagreement}{segmentAgreement(s)}
\DeclareMathOperator{\agreement}{agreement(t)}
\DeclareMathOperator{\predLabels}{predLabels(t)}
\DeclareMathOperator{\goldLabels}{goldLabels(t)}

\begin{document}

\title[Semantic Norm Recognition and its application to Portuguese Law]{Semantic Norm Recognition and its application to Portuguese Law}


\author*[1]{\fnm{Maria} \sur{Duarte}}\email{maria.a.duarte@tecnico.ulisboa.pt}

\author*[1]{\fnm{Pedro A.} \sur{Santos}}\email{pedro.santos@tecnico.ulisboa.pt}
\equalcont{These authors contributed equally to this work.}

\author*[2]{\fnm{João} \sur{Dias}}\email{jmdias@ualg.pt}
\equalcont{These authors contributed equally to this work.}

\author*[3]{\fnm{Jorge} \sur{Baptista}}\email{jorge.manuel.baptista@gmail.com}
\equalcont{These authors contributed equally to this work.}


\affil*[1]{\orgdiv{University of Lisbon, Instituto Superior Técnico}, \orgaddress{\city{Lisbon}, \country{Portugal}}}
\affil*[2]{\orgdiv{Faculty of Science and Technology, University of Algarve and CCMAR and INESC-ID}}
\affil*[3]{\orgdiv{University of Algarve  and INESC-ID}}





\abstract{ Being able to clearly interpret legal texts and fully understanding our rights, obligations and other legal norms has become progressively more important in the digital society. However, simply giving citizens access to the laws is not enough, as there is a need to provide meaningful information that cater to their specific queries and needs. For this, it is necessary to extract the relevant semantic information present in legal texts.
Thus, we introduce the SNR (Semantic Norm Recognition) system, an automatic semantic information extraction system trained on a domain-specific (legal) text corpus taken from Portuguese Consumer Law.  The SNR system uses the Portuguese Bert (BERTimbau) and was trained on a legislative Portuguese corpus. We demonstrate how our system achieved good results (81.44\% F1-score) on this domain-specific corpus, despite existing noise, and how it can be used to improve downstream tasks such as information retrieval.}




\maketitle

\section{Introduction}\label{sec1}



Nowadays many countries have their laws available online. For example, all the norms and laws of the Portuguese Republic, are published by The Official Portuguese Gazette (Diário da República), available online at DRE.pt, under the curation of a state-owned company, INCM (Imprensa Nacional Casa da Moeda). It works as a public service with universal and free access to its content and functionalities. The current search process used enables regular citizens to search for legislation, but it is based on traditional keyword matching algorithms \cite{}. 

The appearance of transformer-based language models has shifted the Natural Language Processing (NLP) area towards semantic models such as Sentence-BERT\cite{}, capable of determining if two sentences have a similar meaning even if they do not share the same words. However, the problem with these models is that they require a large dataset of examples for the models to be fine-tuned to a particular domain, which might not be possible in a service such as DRE, where every month more than a thousand new digital juridical texts are published.



One example of an approach in NLP to automatically extract semantic information from legal texts is the work of Humphreys et al.~\citeyearpar{humphreys2020populating}. The authors extracted definitions and deontic concepts (such as rights and obligations) from legislation, and showed that capturing these legal concepts, and the relationships between the terms that represent such concepts, helps information retrieval. Similarly to Humphrey et al., we hypothesize that the extraction of definitions, norms and semantic relations from legal articles would be beneficial to downstream tasks. By extracting all this information, laws would be represented in a formalism that could then be used to infer knowledge and help build and train NLP-based systems, such as information retrieval and question answering systems. These systems would help citizens to better understand and navigate the legal domain. Thus, the main research questions this article aims to answer are the following:
\begin{itemize}
    \item RQ1: How to create a system capable of automatically extracting and representing norms, entities and semantic relations from Portuguese legal texts?
    \item RQ2: Can the automatic extraction of norms, entities and semantic relations from text improve downstream tasks such as information retrieval?
\end{itemize}

This work focuses on the extraction of several concepts from the Portuguese Legislation. We will focus on a particular subset, the Portuguese Consumer Legislation.  
These concepts include legal norm types (like definitions, rights, obligations, and so on), several types of named entities (like legal references or administrative bodies), and some semantic roles (such as who is obligated, what is that person obligated to, among others), that are defined within those categories. Some of this information is similar to that given in a Semantic Role Labelling (SRL) task, which gives us information regarding ``who" did ``what” to ``whom”, ``when” and ``where”. Additionally, just like Named-Entity recognition (NER) extracts named entities, we want to extract norm types, named entities and semantic roles. Thus, we assume that we were dealing with a NER classification task. Also, having in mind that some of the concepts we want to extract can be nested within each other, i.e. norm types include both entities and semantic roles, we assume we were dealing with a nested NER task. 
Thus, since the type of information we are extracting involves not only semantic roles (as in SRL), but also named entities (as in NER), we name this domain-specific task as Semantic Norm Recognition (SNR).

\section{Related Work}

The NER task can be defined as the extraction and classification of two structural types of entities: 
flat entities and nested entities. 
Most existing NER models focused only on flat NER. 
Some of these works involve applying linear-chain conditional random fields~\cite{lafferty2001conditional} and semi-Markov conditional random fields~\cite{sarawagi2004semi}. 

Still, there are a few models that deal with extracting nested entities, along with flat entities. 
Lu and Roth~\citeyearpar{lu2015joint} proposed a hypergraph model to detect nested entities, which had a linear time complexity. 
Muis and Lu~\citeyearpar{muis2018labeling} improved Lu and Roth model by proposing a multigraph representation based on mention delimiters. They assigned tags between each pair of consecutive words, preventing the model from learning fraudulent structures. These correspond to entities that overlap each other and that are grammatically impossible. 

In the last years, we saw an increasing number of neural models targeting nested NER as well. 
Katiyar and Cardie~\citeyearpar{katiyar2018nested} further extended Muis and Lu approach, by creating a model that learns the hypergraph representation directly for nested entities. This is done by utilizing features extracted from an adaptation of a Long Short Term Memory Neural Network~(LSTM). 
Fisher and Vlachos~\citeyearpar{fisher2019merge} adopted an approach that decomposes nested NER in two stages. Firstly, the boundaries of the named entities at all levels of nesting are identified. Secondly, based on the resulting structure, embeddings for each entity are generated by merging the embeddings of smaller entities/tokens from previous levels. 

Yu, Bohnet and Poesio~\citeyearpar{yu2020named} proposed a model based on the dependency parsing model of Dozat and Manning~\citeyearpar{dozat2016deep} to provide a global view on the input via a biaffine model. Not only did they used a Bidirectional Long Short Term Memory Neural Network (BiLSTM) to learn word representations, they also used two Feed Forward Neural Networks (FFNNs) to generate representations for possible start and end of spans, which they then classify with the biaffine classifier. The authors argued that using different representations for the start and end of spans, allowing the model to learn these representations separately, resulted in an increase in accuracy, when compared to previous systems that did not adopted this idea. They claim that the increase in accuracy comes from the fact that the start and end of spans have different context, thus their learning should be done separately. The authors also showed how their model outperformed the previous ones, mentioned above, for the ACE 2005 Dataset. They also evaluated their model on other seven datasets including ACE2004, GENIA, and ONTONOTES, among others. For all these datasets, their model achieved the best results.

Humphreys et al.~\citeyearpar{humphreys2020populating}, tackled the problem of automating knowledge extraction from legal text, and how to use such knowledge to populate legal ontologies. 
They built a system based on NLP techniques and post-processing rules, derived from domain-specific knowledge. 
This system can be divided into two main components. 
The first is a Mate Tools semantic role labeler~\citep{bjorkelund-etal-2009-multilingual}. This component is responsible for extracting an abstract semantic representation, along with dependency parse trees. The second is a set of rules, responsible for recognizing norms and definitions, classifying their norm type and mapping arguments in the semantic role tree to domain-specific slots in the legal ontology. 

Additionally, and very recently, Neda and Mark~\citeyearpar{DBLP:journals/corr/abs-1812-01567} proposed an information extraction process to build a legislation network. Due to the structured nature of their legal corpus they were able to apply NER by defining clear rules to extract the relevant entities.

Finally, Ruggeri et at.~\citeyearpar{article} proposed a data-driven approach for detecting and explaining fairness in legal contracts for consumers. They applied a Memory-Augmented Neural Network~(MANN, ~\cite{articleMANN}) in order to incorporate external knowledge (set of legal explanations) when classifying legal clauses.

\section{Dataset}\label{sec8}
As mentioned above, the goal of this work is to create an automatic information extraction system, named SNR, capable of extracting relevant concepts (such as norms, semantic roles and named entities), from a specific corpus that contains the Portuguese Consumer Law. 
Since we are dealing with a classification problem, we need to have a dataset with the laws and corresponding gold labels. 
For this reason, the corpus was annotated so we could generate the needed dataset for our task. 
This section covers the creation of the dataset, with a brief description of the corpus, of the annotation process and the tools used.

\subsection{Corpus}

The SNR system uses a subset of legal articles from the Portuguese Consumer Legislation.
These legal texts were made available by INCM and were segmented at the level of the article number, containing in total 5.600 segments, and 341.392 tokens. This segmentation was done at the level of the article number, due to the fact of this being considered the level that gave the necessary context to extract and annotate the relevant semantic information, as an article number often corresponds to a full sentence. 

Thus, each segment, roughly speaking, corresponds to a number of an article of a legal act. 
For example, 
\say{1 - É revogada a Lei n.º 29/81, de 22 de agosto.} 
(1 - The law n.º 29/81, from August 22, is revoked.),
is the segment with ID 5839, and corresponds to the number 1 of article 24 of the Decree-Law no. 47/2014 dated from the 28th of July.


Each segment is associated with an unique identifier (segment ID) and other metadata regarding the legal text from which the segment was extracted from. This information includes the number of the legal act, its publishing date, chapter, paragraph number, and so on.

\subsection{Annotation Procedure}
In the context of this work, the goal of the SNR system is to extract the norms' types, the semantic roles and the named entities present in the legal text. 
Thus, the corpus must be annotated with these concepts. 
The relevant norm types chosen were based on the concepts mentioned in Humphreys et al. \citeyearpar{humphreys2020populating}, and considering a preliminary analysis of the corpus.
They consist in norms conveying the deontic constructs of obligations (\textsc{oblig}) and rights (\textsc{right}); articles presenting legal definitions (\textsc{def}); statements about the entry into force or the revocation of specific laws (\textsc{leffect}); and residual introductory articles (\textsc{intro}), framing the context, or stating the purpose or the domain of application of the current legal act.

Named entities (NE) include domain-specific NE, such as legal document references (\textsc{lref}), like the law mentioned in the example above, and other textual references (\textsc{tref}), often referring to other articles/sections/items within the current legal act. The later require anaphora resolution (not considered at this stage). 
Other, more typical, NE were also tagged, such as time expressions like those denoting dates (\textsc{time\_date}), duration (\textsc{time\_duration}) and frequency (\textsc{time\_freq}). Distinction between absolute (\textsc{time\_date\_abs}) and relative dates was considered, the later requiring some temporal anaphora resolution (not considered at this stage). 
The relative dates may refer either to the moment of utterance (=date of publication), which is rare, or to an event already mentioned in the text (\textsc{time\_date\_rel\_text}). 
Considering their frequency, only this later type and the duration type were included in the model.
The names of administrative bodies (\textsc{ne\_adm}) and other organizations (\textsc{ne\_org}) are also considered, including office titles (\textsc{ne\_office}). 

As for the semantic roles (SR) included in the task, some are specific of a given norm type, such as those involved in definitions, v.g. \textsc{definiendum}, for the concept to be defined; \textsc{definiens}, for the definition proper; 
and \textsc{scope}, delimiting the domain of application of a definition.
Other SR denote commonly occurring circumstantial events, such as \textsc{condition}, \textsc{purpose}, \textsc{concession}, \textsc{exception}. 
For the \textsc{right} and \textsc{oblig} norm types, which may arguably be considered the most important concepts to be captured from the legal texts, specific roles were defined for the \textsc{experiencer}, that is, the person/entity holding the right or having an obligation; and the \textsc{action}, i.e. the very content of that right and/or that obligation. 
Finally, negation expressions (\textsc{neg}), which reverse the polarity of the deontic value attributed to the \textsc{action} by either the obligation (\textsc{oblig}) or the \textsc{right} norm type, were explicitly annotated.
The list of all chosen concepts can be seen in Table~\ref{table:label_table}, and were annotated using the Prodigy tool\footnote{https://prodi.gy/}.


\begin{table}[h!]
\centering
\caption{All 23 chosen concepts}
\begin{tabular}{|c|} 
 \hline
 \textbf{Norms} \\ 
 \hline
 DEF \\
 INTRO \\ 
 LEFFECT \\
 OBLIG \\
 RIGHT \\
 \hline
\end{tabular}
 \quad
 \begin{tabular}{|c|} 
 \hline
 \textbf{Named Entities} \\ 
 \hline
 LREF \\
 TREF \\
 NE\_ADM \\
 TIME\_DURATION \\
 TIME\_DATE\_REL\_TEXT \\ 
 \hline
\end{tabular}
 \quad
 \begin{tabular}{|c|} 
 \hline
 \textbf{Semantic Role}s \\ 
 \hline
 ACTION \\ 
 CONCESSION \\
 CONDITION \\
 DEFINIENDUM \\
 DEFINIENS \\
 DEF-INCLUSION \\
 EFFECT \\
 EXCEPTION \\
 EXPERIENCER \\
 NEG \\
 PURPOSE \\
 SCOPE \\
 THEME \\
 \hline
\end{tabular}
\label{table:label_table}
\end{table}

A set of guidelines was produced to precisely define and abundantly exemplify all the relevant concepts here involved, and to provide specific orientations for more complex or problematic cases. 
The annotation was then carried out by a team of linguists, following these guidelines. 
This process was developed in two main steps. On a first step, each segment as a whole was tagged with its corresponding norm type. Then, on the more granular second step, the named entities within each segment were delimited and classified. Finally, the semantic roles were identified and tagged. 
At each step, a pilot-annotation was carried out, to train the annotators and assess the difficulties posed by the texts, or by inconsistencies and lacunae found in the guidelines, which have accordingly been revised. 
In the end, inter-annotator agreement as calculated. An average Fleiss Kappa coefficient of 0.79 was found for the first step (norm type classification), which can be interpreted as ``substantial'' or ``strong''. 
For the second step, the task was conceptually much more complex and, in spite of the guidelines, several inconsistencies were found among the annotators, especially for some semantic relations and for some complex time-related named entities. Also, some semantic roles had been missed.
These situations introduced some noise in the dataset.
Systematic revision was then undertaken to correct the more noisy labels and to improve the overall quality of the annotation. 
For example, several \textsc{def-inclusion} spans, which correspond basically to an enumeration of items within a definition, had been incorrectly assigned to alternative definitions (\textsc{definiens}) of the same concept (\textsc{definiendum}); the opposite case was also found in the dataset. 
In other cases, several lexical cues and some patterns of co-occurrent semantic roles were investigated and, then, missing or inconsistent annotations were corrected. 
For example, several instances of compound conjunction``desde que'' ('as long as') introducing a \textsc{condition} SR span had been missed, hence precluding the identification of those spans. 
Patterns like these were systematically revised and corrected.
As a result of this correction phase, the dataset went from an initial number of 34,178 marked spans to a final total of 36,711.

\section{Semantic Norm Recognition System}\label{sec9}
In this section we will describe our SNR system and all the implemented approaches.

\subsection{Baseline SNR System}
The purpose of this work is to create a system capable of identifying norm types, semantic roles and named entities present in legal texts. 
These concepts can appear nested, which means, for example, there can be sentences where an \textsc{experiencer} is inside an \textsc{action} (both, semantic roles). 
Once the annotation procedure was done, we saw that we could have more than one label for the same span, with each label belonging to a different group of spans (either a norm, or a semantic role or a named entity). 
For example, a certain span could not only be an \textsc{experiencer}~(semantic role), but also a \textsc{ne\_adm}~(named entity). 
These, and only these cases, correspond to a multi-label problem. A span that is labeled with a concept of a certain group (either norms, semantic roles or named entities) will never be labeled with another concept of the same group. For example, a certain span that is labeled as \textsc{experiencer}, will not have any other semantic role associated with it. Thus, for each group we have a multi-class problem.

\begin{figure}[H]
 \centering
 \includegraphics[width=10cm]{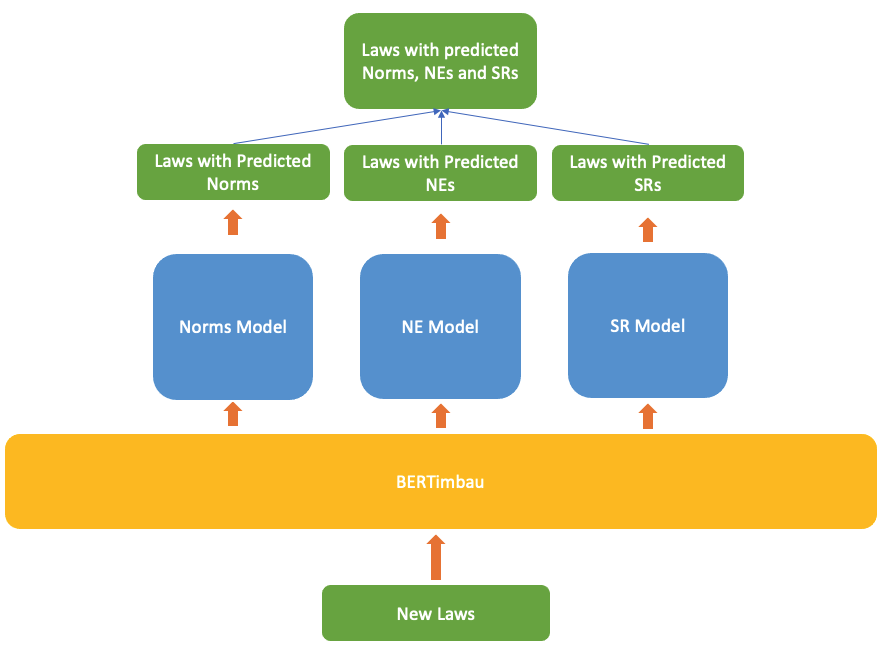}
 \caption{Baseline SNR System}
 \label{fig:snr_phase1}
\end{figure}

In order to make sure that we were dealing only with a multi-class problem for all concepts and not a multi-label problem, we created an information extraction system composed of three models: 
(i) the Norms Model, responsible for predicting the norms; 
(ii) the Named Entities~(NE) Model, responsible for predicting the named entities; and 
(iii) the Semantic Roles~(SR) Model, responsible for predicting the semantic roles. 
All models have the same architecture, but are trained to learn different types of labels (norms, named entities and semantic roles, respectively). 
The overview of the SNR system is shown in Figure \ref{fig:snr_phase1}. 

\begin{figure}[H]
 \centering
 \includegraphics[width=11cm]{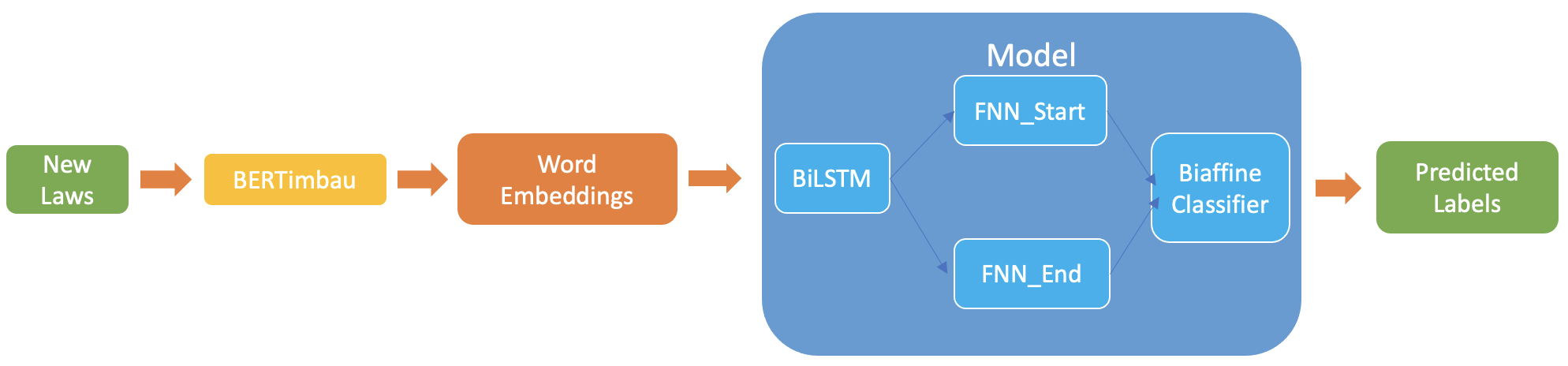}
 \caption{Architecture of each SNR System Model}
 \label{fig:snr_model}
\end{figure}

The architecture of each model is based on the dependency parsing model of Yu, Bohnet and Poesioo~\citeyearpar{yu2020named}, with a small difference regarding the embeddings used, as we can see in Figure \ref{fig:snr_model}. 

\noindent These authors used BERT, fastText and charCNN embeddings. Due to the nature of the corpus, that consists of Portuguese legal text, we used Portuguese BERT (BERTimbau, ~\cite{souza2020bertimbau}) as word embeddings. 
BERTimbau is the Portuguese version of BERT. 
As BERT is the state-of-the-art for word embeddings, we decided there was no need to use fastText. Regarding charCNN, these characters embeddings are mostly used to deal with emoticons and misspelling words. 
Since we are dealing with legislative text, which is subject to intense scrutiny and careful edition prior to its publication, even the existence of misspelled words will be rare. Therefore, this type of embeddings was not included in our system. 

Similarly to \cite{yu2020named}, we fed our embeddings into a BiLSTM, to learn the words representation, and applied two FFNNs to the word representations generated by the BiLSTM, in order to learn different representations for the start and the end of the spans. Each FFNN computed representations $h_s$ and $h_e$, for the start and the end of the entities, respectively. Then, we applied a biaffine classifier~(\cite{dozat2016deep}) in order to generate a scoring tensor $r$, of size $l \times l \times c$, over the input sentence, where $l$ corresponds to the sentence length and $c$ to the number of entities (norms, for the Norms Model; semantic roles, for the Semantic Roles (SR) Model; and named entities, for the Named Entities (NE) Model), plus one to represent non-entities. The score for each span $i$, corresponds to:

{\centering \[r(i) = h_s(i)^T U h_e(i) + W(h_s(i) \bigoplus h_e(i)) + b\] \par }

\noindent where $U$ is a $d \times c \times d$  tensor, $W$ a $2d \times c$ matrix and $b$ the bias. All the valid spans (spans whose end is after its start) are scored by the tensor. Then, the entity label with the highest score is  assigned for each span:

{\centering \[y'(i) = argmax \; r(i)\] \par }

After having all spans and their possible labels, just like Yu et al.~\citeyearpar{yu2020named}, we also did a final post-processing step to make sure that there were no nested entities whose boundaries clash. 
For example, the spans ``the article number 5 defines an" and ``article number 5 defines an obligation to consumers" clash with each other (boundaries overlap), and so in this situation the selected span would be the one with a higher score. 
The model's learning goal is to assign the right category (including non-entity) to each valid span. Being a multi-class classification problem, we optimised the model with softmax cross-entropy.



 
\subsection{Full Norm Dependency SNR System}

\begin{figure}[H]
 \centering
 \includegraphics[width=10cm]{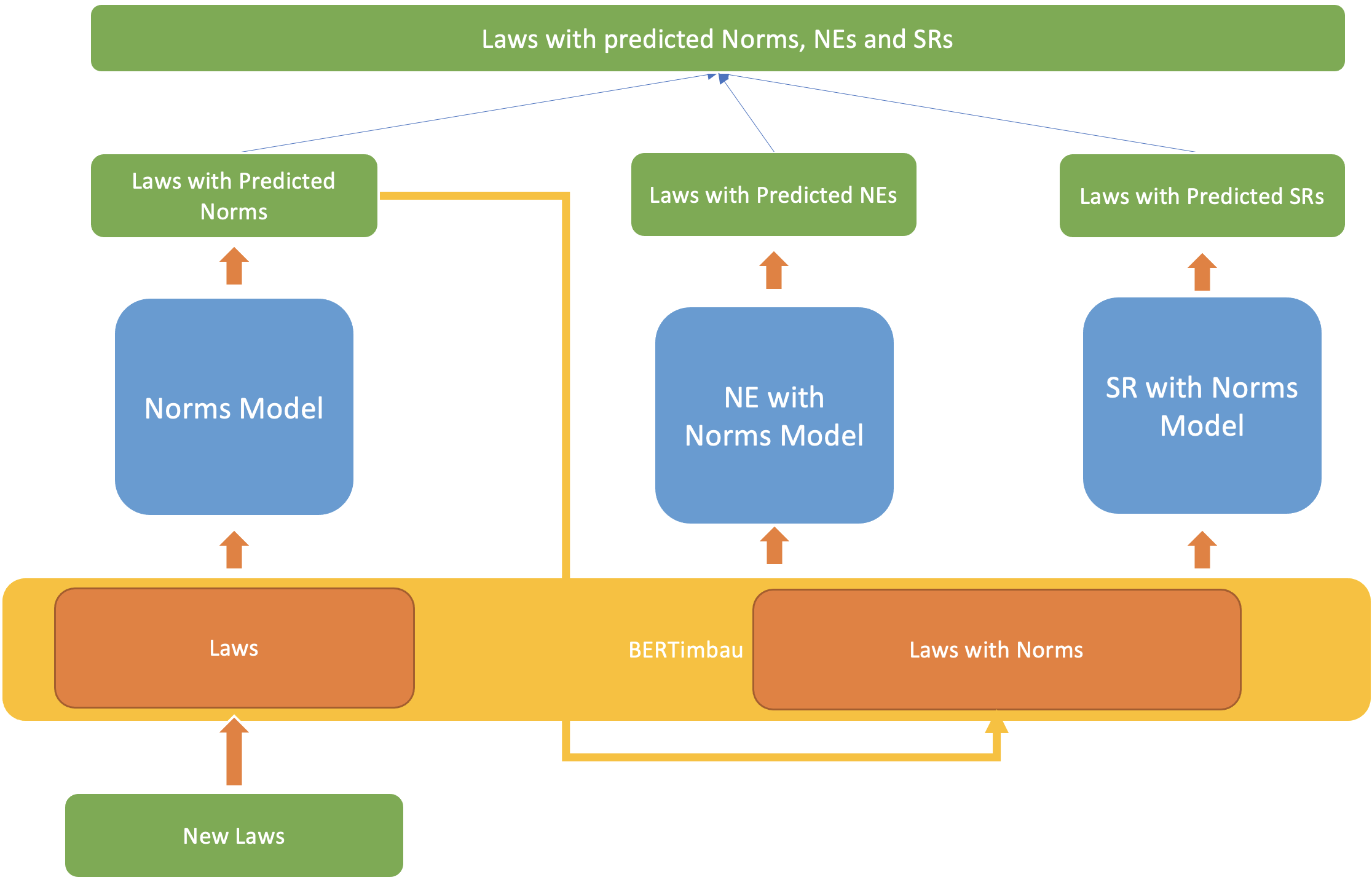}
 \caption{Full Norm Dependency SNR System}
 \label{fig:snr_phase2}
\end{figure}

In the previous section we described the  \emph{Baseline SNR system}. 
It was built in order to learn to predict all the information that was annotated. 
However, the annotation itself was divided into two levels: first, identifying the norm types for the whole corpus; and, then, for those norms, identifying the remaining concepts. 
We, thus, decided to create another version of the system, which we denote by \emph{Full Norm Dependency SNR System}, as shown in Figure \ref{fig:snr_phase2}, in order to try to replicate the annotation process and the human rationale behind it.

The Full Norm Dependency system is very similar to the Baseline system, the key difference being the fact that the system tries to learn to identify the named entities and semantic roles, knowing the norms that are present in those segments. 
The rationale behind this strategy is that by predicting the norm type first (first phase), it would allow the system to focus on predicting the semantic roles and named entities for that particular norm, making this second phase easier. 
This was also the strategy adopted in the (human) annotation process.

This system starts by generating the word embeddings using BERTimbau, just like the Baseline System. 
Afterwards, it uses those embeddings as input for the Norms Model. 
The model will, then, make its predictions regarding the norms, returning all the predicted norms. 
After this, these predicted norms will be added into the segments and fed into our feature extractor (BERTimbau) to generate another set of embeddings. 
For example, for the segment 
``\textit{1
 - É revogada a lei ... no dia anterior.}'' 
 (The law is revoked ... on the previous day),
if the Norms Model predicted the norm type  \textsc{leffect} spanning from token 2 (``\textit{É}'') to token 22 (``\textit{anterior}''), instead of feeding BERTimbau with the original segment, we feed it with ``\textit{1 -} \textsc{ileffect} \textit{É revogada a lei ... no dia anterior} \textsc{fleffect}''. 
In short, we added \textsc{ILabel} and and \textsc{FLabel} to represent the start and the end of the norm, respectively, with \textsc{Label} being the label of the corresponding norm. 
After generating these embeddings, we feed them to two models responsible for predicting the named entities and the semantic roles, respectively. 
These two new models differ from the Named Entities and Semantic Roles Model of the baseline approach, since they are trained with more information (the norm types) than those two models from the baseline, which had only been trained with the original segments. 
Let us denote these new two models by \textit{Named Entities Norm Dependent} (NEND) \textit{Model}, and \textit{Semantic Roles Norm Dependent} (SRND) \textit{Model}, respectively. 
Finally, like the Baseline system, the predictions of each model are concatenated together, in order to obtain the final classification.
 
As we mentioned before, the system uses the predicted norms as input for the other two models. 
Regarding the training of the two models, however, we did not feed the models with the predicted norms in the training and validation segments. 
Instead, we used the gold labels in order to have each model learn correctly the relations between the norms and other labels (named entities for the NEND Model, and semantic roles for the SRND Model), as we do not want the model to learn wrong relations. 
For example, let us consider that we have a segment that has a \textsc{def}~(norm type) and also has a  \textsc{definiendum}~(semantic role). 
If we were training the SRND Model, and if the Norms Model had incorrectly predicted (assuming it had been trained already) that the segment would have been tagged as an \textsc{oblig} instead of a \textsc{def} norm type, by using the incorrect label, the model could learn to associate \textsc{oblig} with \textsc{definiendum}. 
Therefore, to make sure that the model only learned the correct relations, we only used the gold labels. 
This also allows us to train all three models simultaneously, after generating the two sets of embeddings (one for the original segments, and another for the segments with the norms gold labels) .

\subsection{Partial  Norm Dependency SNR System}

\begin{figure}[H]
 \centering
 \includegraphics[width=10cm]{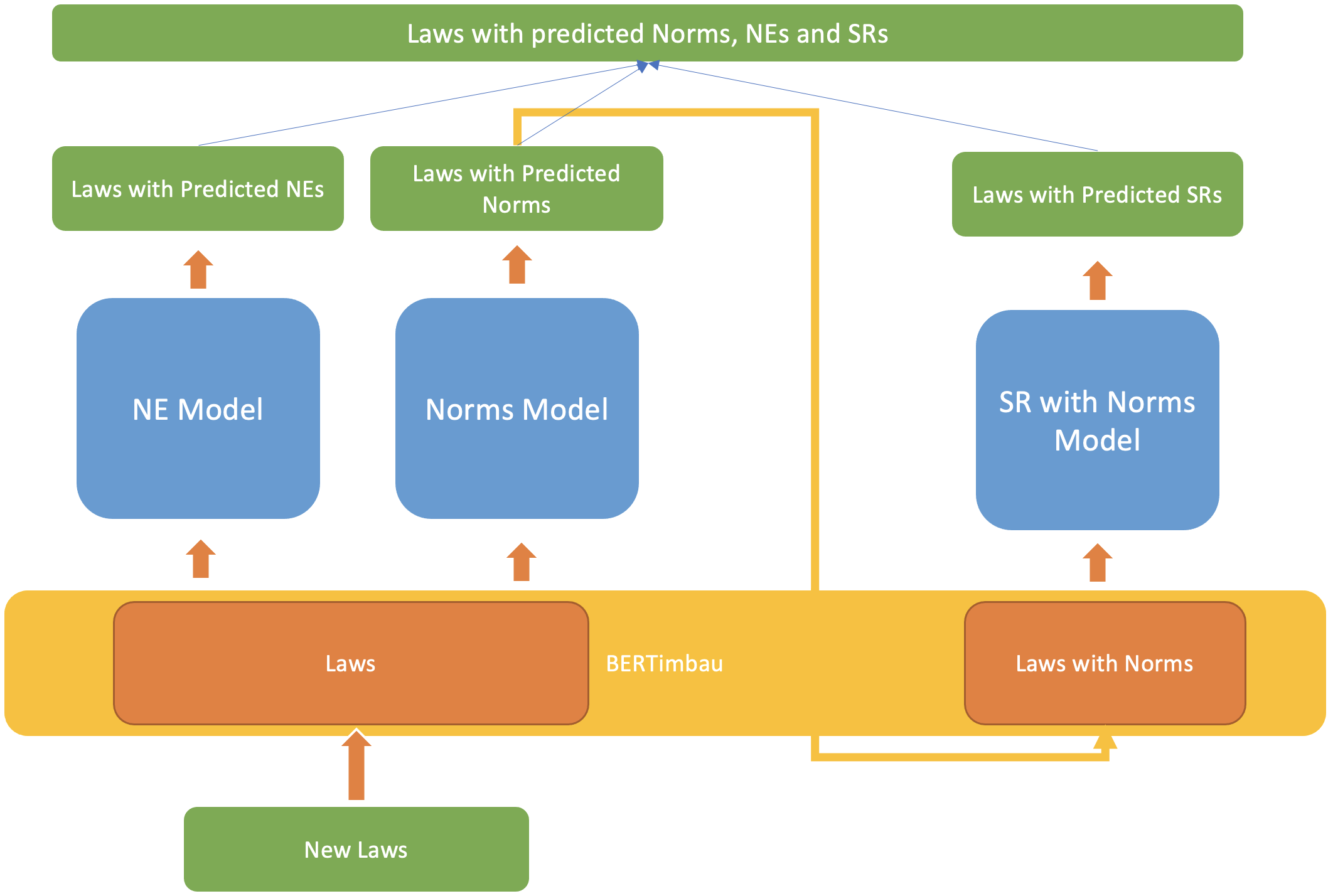}
 \caption{Partial Norm Dependency SNR System}
 \label{fig:snr_Partial  Norm Dependency}
\end{figure}

After doing some experimental training and evaluating with both of our previous systems, we saw that some named entities seemed to have worse results when using the Full Norm Dependency system. After looking at each named entity and semantic role, we saw there was a stronger relationship between norms and semantic roles than there was between norms and entities. For example, in the corpus \textsc{lref} or \textsc{tref} are text-related named entities that can occur in any type of norm, but we can only have a \textsc{definiendum} inside a \textsc{def}, or a \textsc{effect} inside a \textsc{leffect}. Having this in mind, we decided to create a third version of our system, which we denoted as \textit{Partial  Norm Dependency SNR} System, which is shown in Figure \ref{fig:snr_Partial  Norm Dependency}. 

As we can see, the Partial Norm Dependency approach is very similar to the Full Norm Dependency. In the previous approach, we had the NEND Model and SRND Model responsible for predicting the corresponding labels, knowing the norms that were present in the segments. In this approach, instead, we only use the norms' information for the model responsible for predicting the semantic roles (SRND Model). Thus, we use the Named Entities Model, instead of the NEND Model to predict the named entities.

\begin{figure}[H]
 \centering
 \includegraphics[width=12cm]{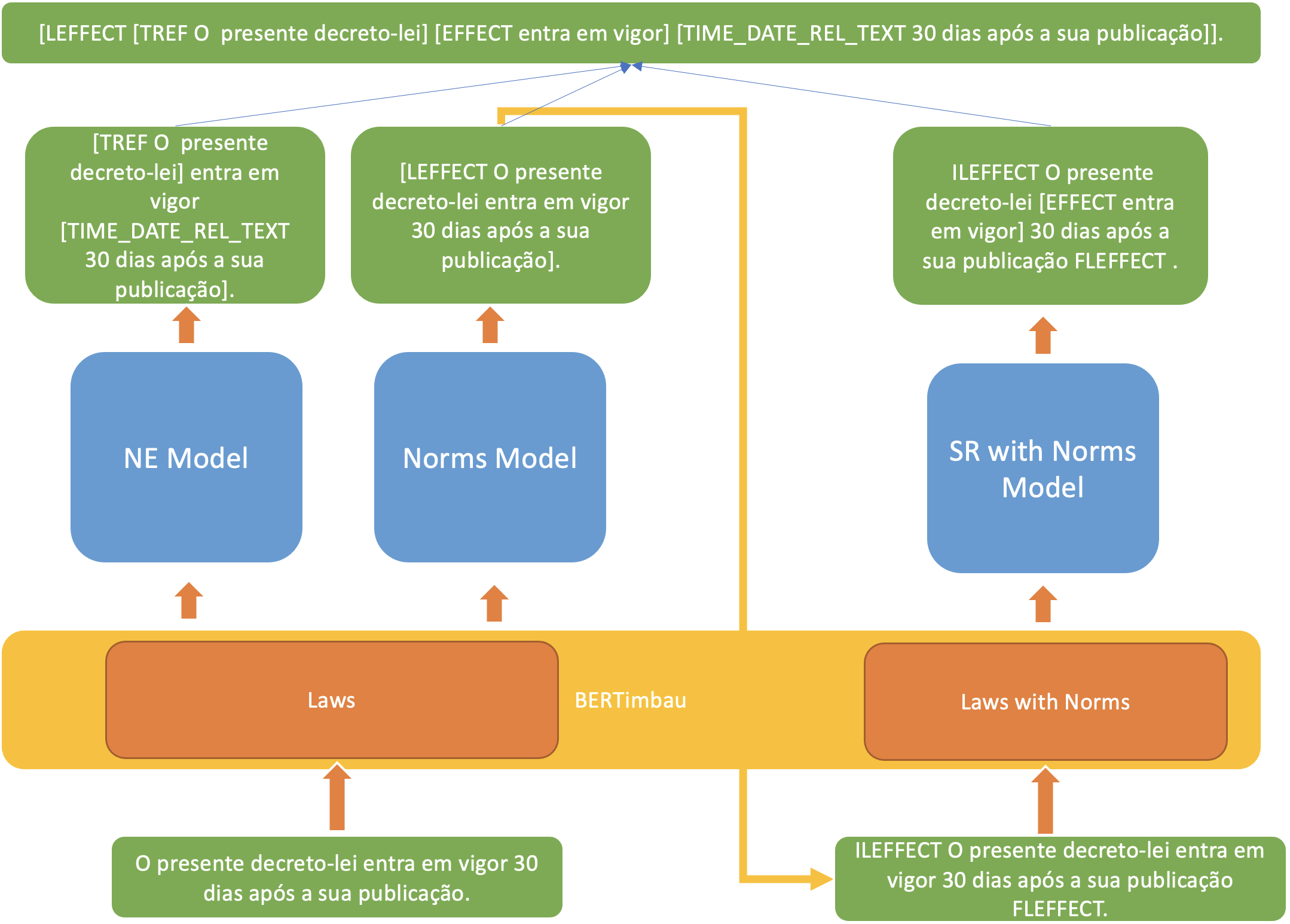}
 \caption{Partial  Norm Dependency SNR System Input/Output Example}
 \label{fig:snr_Partial  Norm Dependency_ex}
\end{figure}

To get a better understanding of the input and output of our system, see Figure \ref{fig:snr_Partial  Norm Dependency_ex}, which includes the predictions the Partial Norm Dependency SNR System makes for the input segment ``\textit{O presente decreto-lei entra em vigor 30 após a sua publicação.}” (The current decree-law come into effect 30 days after its publication).

\section{SNR Training, Optimization and Evaluation}\label{sec10}

\subsection{Models Training and Optimization}
As we mentioned above, we created three different versions of our system. 
In total there are five different models: the Norms Model, the NE Model, the SR Model, the NEND Model, and, finally, the SRND Model. 
We applied a division of {80\% training}/{10\% validation}/{10\% test} to the dataset, used batches of size 32, trained all models for 15 maximum epochs (saving the one with highest F1-score), and optimized the parameters using Bayesian optimization. 
The results can be seen in Table \ref{table:optimizationResultsfinal}.

\begin{table}[h!]
\centering
\caption{Partial  Norm Dependency SNR System: Norms Model Results}
\begin{tabular}{|c c c c c c|} 
 \hline
 Model & Learning Rate & Decay Rate & FFNN Size & BILSTM Size & F1-score \\ 
 \hline
 Norms & 0.004 & 0.92 & 39 & 410 & 84.38\% \\
 NE & 0.004 & 0.86 & 75 & 252 & 86.76\% \\
 SR & 0.008 & 0.93 & 152 & 218 & 74.89\% \\
 NE With Norms & 0.005 & 0.87 & 54 & 284 & 86.17\% \\
 SR With Norms & 0.003 & 0.88 & 134 & 273 & 77.35\% \\ 
 \hline
\end{tabular}
\label{table:optimizationResultsfinal}
\end{table}



\subsection{Evaluation Metrics}
For evaluating the system and each model, we used Micro-F1, Micro-Precision and Micro-Recall. 
To evaluate each label, we used the standard F1-score, Precision and Recall.

We also created a novel metric, which we named  ``Average Token Agreement ($ATA$)''. Since the previous metrics only count exact matches as correct, we decided it would be useful to know how many tokens we correctly predicted. This way, we would have information regarding the agreement between the gold labels and predicted labels at token level, and not just whether it is a exact match or not. The $ATA$ score of a set $S$ of segments can be calculated by 


\begin{equation}
    \label{AKT3}
 \ata = \frac{ \sum_{s\in S}{ \segmentagreement}}{\vert S\vert },
\end{equation}
where
\begin{align*}
  \segmentagreement &= \frac{\sum_{t}{\agreement}}{\vert s\vert},\\[3mm]
 \agreement &= \frac{ \mid \goldLabels \cap \predLabels \mid}{ \mid \goldLabels \cup \predLabels \mid},
\end{align*}

\noindent with $\vert S\vert$ corresponding to the number of segments in the set, $\vert s \vert$ to the number of tokens in the segment, $t$ to a token of the segment $s$,  $\goldLabels$ to the gold labels, and $\predLabels$ to the predicted labels, of the token $t$.

\begin{figure}[H]
 \centering
 \includegraphics[width=9cm]{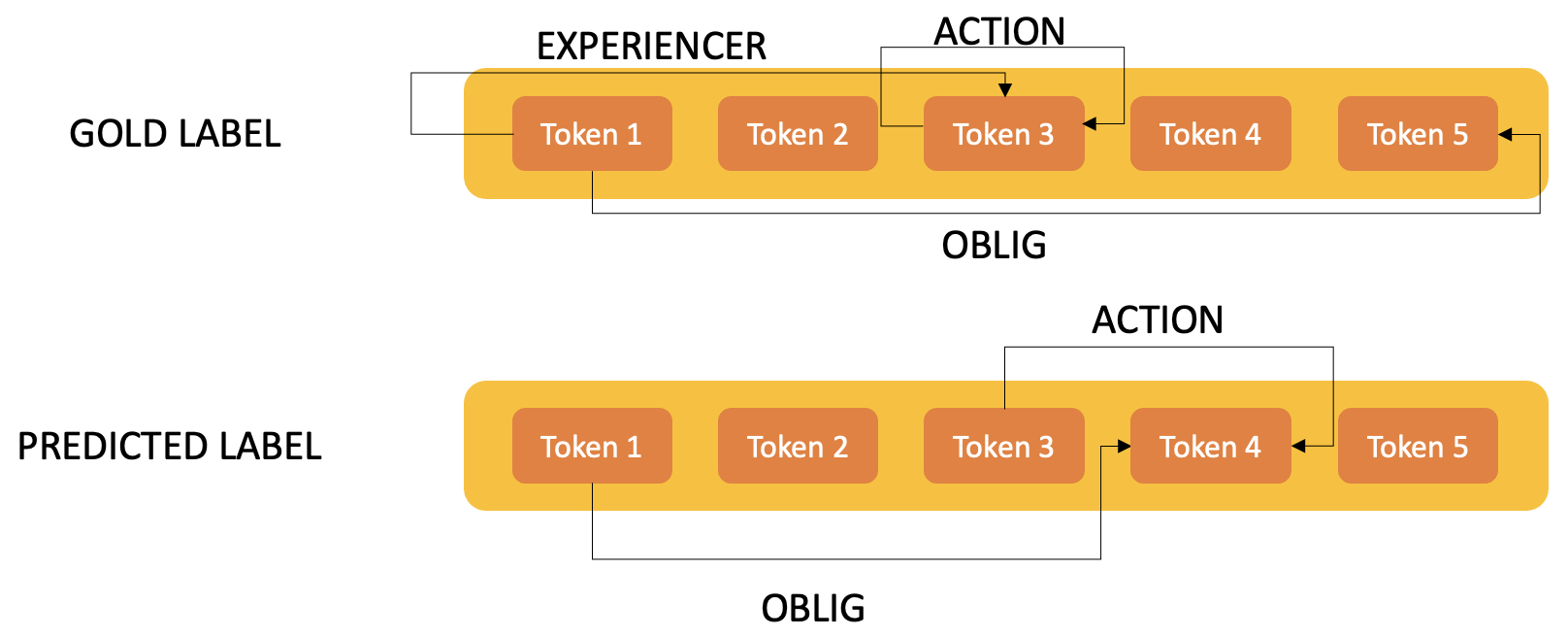}
 \caption{Example for agreement metric}
 \label{fig:agreement}
\end{figure}

Figure \ref{fig:agreement} shows a fictitious  segment and its corresponding gold and predicted labels for the purpose of illustrating the $\ata$ metric.
Each arrow represents a span and its legend corresponds to its gold/predicted label (semantic roles above and norm types below). 
The $\ata$ given by the system for that single segment, is (1/2 + 1/2 + 2/3 + 1/2 + 0/1)/5 = 0.43. 
For example, for Token 1, the gold and predicted labels agree in 1 label (\textsc{oblig}) over 2 labels (\textsc{oblig} and \textsc{experiencer}). 
In this case, 0.43 means that the system overall and the gold labels agree with each other 43\%, regarding that sentence. For x sentences, the agreement would be the sum of the agreement of each sentence divided by the number of sentences, as we can see in the Equation \eqref{AKT3}.

If we consider a specific model, and not the full system, for example, the SR Model, the agreement for this same sentence would be (0/1 + 0/1 + 1/2 + 0/1 + 1)/5 = 0.3, which means the gold and predicted labels have an agreement of 30\%, regarding the semantic roles, which makes sense, since from 5 tokens in total, only in Token 3 and in Token 5 do the predicted and gold labels match (in case of Token 3 only in 1 of 2 of its semantic roles). Also as we can see tokens that have no labels (in this case Token 5 has no semantic role) will not be considered.

\subsection{Results' Comparison}
After performing evaluation for all three approaches, we achieved the results shown in Table \ref{table:1}.

\begin{table}[h!]
\centering
\caption{Performance of each model and each SNR System approach}
\begin{tabular}{|c c c c c|} 
 \hline
 Model & F1-score & Recall & Precision & $ATA$ \\ 
 \hline
 Norms & 84.74\% & 83.01\% & 86.54\% & 0.91 \\
 NE & 88.67\% & 90.45\% & 86.96\% & 0.98 \\
 SR & 76.06\% & 73.12\% & 79.25\% & 0.81 \\
 NEND & 86.94\% & 84.52\% & 86.54\% & 0.98 \\
 SRND & 76.86\% & 78.72\% & 86.54\% & 0.83 \\ 
 \hline\hline
 SNR System & F1-score & Recall & Precision & $ATA$ \\ 
 \hline
 Baseline SNR & 81.05\% & 79.41\% & 82.75\% & 0.85 \\
 Full Norm Dependency SNR & 81.00\% & 80.26\% & 81.75\% & 0.84 \\
 Partial  Norm Dependency SNR & 81.44\% & 80.50\% & 82.40\% & 0.85 \\ 
 \hline
\end{tabular}
\label{table:1}
\end{table}


Immediately by looking at the results, we can see that the NE Model outperforms the NEND Model in all metrics (F1-score, Precision and Recall) except the Agreement, which is very close. Thus, the Full Norm Dependency approach is not the best approach. The SRND Model has a higher F1-score than the SR Model. This comes from the fact that, even though it has a smaller precision than the SR Model, the SRND Model has a higher recall (being a bigger difference when comparing increase in precision) than the SR Model. This is reflected on the results, when we compare the Baseline Approach with the Partial  Norm Dependency approach and see that the Partial  Norm Dependency approach has a higher F1-score and recall than the Baseline approach. Based on these results and since our priority is the model making the highest number of corrected span predictions (F1-score), we concluded that the best approach is the Partial  Norm Dependency approach. 

Even thought the Partial  Norm Dependency approach is the best approach, this and the Full Norm Dependency approach, have some limitations when compared to the baseline one. The SRND Model and the NEND Model are dependent on the predictions the Norms Model did. The consequence of this is that when the Norms Model makes wrong predictions, it can lead the other two models associating the wrong information with the one it was provided. For example, the Norms Model incorrectly predicted 11 \textsc{def}s to be \textsc{oblig}s. This means that for those segments, the SR with Norms received the segments with the start and end tokens of the predicted \textsc{oblig}s, which could have caused the model to associate labels such as \textsc{action} and \textsc{theme}, instead of \textsc{definiendum} or \textsc{definiens}, leading to wrong predictions. Still, since the Norms Model achieves such good results, this dependency does not seem to have much impact on the performance of these approaches. 



\subsection{Partial  Norm Dependency SNR System Evaluation}
\label{Partial  Norm Dependencyeresults}

Now that we know the Partial Norm Dependency to be the best approach we need to evaluate our system based on its results regarding each label that its trying to predict. Tables \ref{table:Partial  Norm Dependency_norms_results}, \ref{table:Partial  Norm Dependency_ne_results} and \ref{table:Partial  Norm Dependency_sr_results}, show the results of each model of the Partial  Norm Dependency SNR system. 

\begin{table}[h!]
\centering
\caption{Partial  Norm Dependency SNR System: Norms Model Results}
\begin{tabular}{|c c c c c|} 
 \hline
 \multicolumn{5}{|c|}{Norms Model} \\
 \hline
 Labels & F1-score & Recall & Precision & $ATA$ \\
 \hline
 DEF & 68.97\% & 62.5\% & 76.92\% & 0.63 \\
 OBLIG & 88.37\% & 87.44\% & 89.31\% & 0.88 \\
 RIGHT & 74.18\% & 72.48\% & 75.96\% & 0.67 \\
 LEFFECT & 100\% & 100\% & 100\% & 1 \\
 INTRO & 80\% & 66.67\% & 100\% & 0.67 \\ 
 \hline
\end{tabular}
\label{table:Partial  Norm Dependency_norms_results}
\end{table}


From looking at Table \ref{table:Partial  Norm Dependency_norms_results}, the Norms Model seems to have good results for almost every norm (74\% to 100\% F1-score), except for \textsc{def}, which has satisfactory results (68.97\% F1-score). \textsc{leffect} was the norm that had the best results having a 100\% F1-score, precision and recall, as well as an $ATA$ of 1. This type of norm is probably the one which we found to be more consistently annotated since it was probably the most simpler concept to identify. \textsc{def} on the other hand, was a norm which caused some ambiguity, especially in some cases were segments were too long and included many definitions which might have cause indecision of were to start and end each definition, which in turn could explain why the $ATA$ for that norms is so low. 

\begin{table}[h!]
\centering
\caption{Partial  Norm Dependency SNR System: NE Model Results}
\begin{tabular}{|c c c c c|} 
 \hline
 \multicolumn{5}{|c|}{NE Model} \\
 \hline
 Labels & F1-score & Recall & Precision & $ATA$ \\ 
 \hline
 LREF & 90.77\% & 96.72\% & 85.51\% & 0.83 \\
 TREF & 91.76\% & 93.62\% & 89.98\% & 0.92 \\
 NE\_ADM & 89.37\% & 90.61\% & 88.17\% & 0.82 \\
 TIME\_DATE\_REL\_TEXT & 62.63\% & 58.49\% & 67.39\% & 0.52 \\
 TIME\_DURATION & 62.22\% & 63.64\% & 60.87\% & 0.43 \\ 
 \hline
\end{tabular}
\label{table:Partial  Norm Dependency_ne_results}
\end{table}


The NE Model, has great results for every named entity (80\% to 92\% F1-score) except \textsc{time\_duration} and \textsc{time\_date\_rel\_text}, which had satisfactory results. Based on the NE Model predictions we saw that sometimes the model confused these two labels, it could be that since both are temporal expression the model had some difficulty identifying each one separately.

For the SRND Model, only \textsc{definiens}, \textsc{action}, and \textsc{purpose} had satisfactory results, the remaining semantic roles all had pretty good results. \textsc{action} was the semantic role most common in our dataset, which would make us think it should have better results, since it had more samples to train on. Yet, we can see that it had a good $ATA$ score (0.76), which may indicate that some of the predictions that were incorrect were only off a few tokens. From all the spans that corresponded to an \textsc{action}, 182 were predicted to be \textsc{nolabel} by the system, this could contain cases where the predicted span was incorrect by only lacking a certain tokens. 

\begin{table}[h!]
\centering
\caption{Partial  Norm Dependency SNR System: SRND Model Results}
\begin{tabular}{|c c c c c|} 
 \hline
 \multicolumn{5}{|c|}{SRND Model} \\
 \hline
 Labels & F1-score & Recall & Precision & $ATA$ \\ 
 \hline
 DEFINIENDUM & 77.27\% & 66.67\% & 91.89\% & 0.65 \\
 DEFINIENS & 64.71\% & 56.41\% & 75.86\% & 0.46 \\
 DEF-INCLUSION & 83.24\% & 81.91\% & 84.62\% & 0.61 \\
 SCOPE & 72.96\% & 69.05\% & 77.33\% & 0.63 \\
 ACTION & 69.35\% & 65.60\% & 73.56\% & 0.76 \\ 
 CONDITION & 81.45\% & 80.72\% & 82.19\% & 0.78 \\
 CONCESSION & 92\% & 93.88\% & 90.20\% & 0.89 \\
 PURPOSE & 65.85\% & 69.23\% & 62.79\% & 0.51 \\
 EXPERIENCER & 81.28\% & 75.34\% & 88.24\% & 0.78 \\
 THEME & 79.95\% & 83.02\% & 77.09\% & 0.80 \\
 EXCEPTION & 80\% & 82.76\% & 77.42\% & 0.75 \\
 EFFECT & 82.54\% & 86.67\% & 78.79\% & 0.87 \\
 NEG & 86.36\% & 95\% & 79.17\% & 0.76 \\
 \hline
\end{tabular}
\label{table:Partial  Norm Dependency_sr_results}
\end{table}
The mistakes the models made (such as confusing certain norms with each other as well as confusing the two temporal expressions) might come from noise still in the dataset, as we had already found some situations like this, which we talk about in more detail in the following section.

\begin{table}[h!]
\centering
\caption{Partial  Norm Dependency SNR System Results}
\begin{tabular}{|c c c c |} 
\hline
 \multicolumn{4}{|c|}{Partial  Norm Dependency System Performance} \\
 \hline
 F1-score & Recall & Precision & $ATA$ \\ 
 \hline
 81.44\% & 80.50\% & 82.40\% & 0.85 \\ 
 \hline
\end{tabular}

\begin{tabular}{|c c c c c|} 
 \hline
 Model & F1-score & Recall & Precision & $ATA$ \\ 
 \hline
 Norms & 84.74\% & 83.01\% & 86.54\% & 0.91 \\
 NE & 88.67\% & 90.45\% & 86.96\% & 0.98 \\
 SR With Norms & 76.86\% & 75.08\% & 78.72\% & 0.83 \\ 
 \hline
\end{tabular}
\label{table:basleine_results}
\end{table}


Overall, our Partial  Norm Dependency SNR system reached high results, achieving 81.44\% F1-score. The NE Model achieved the best results, with a 88.67\% F1-score, followed by the Norms Model, with a 84.74\% F1-score, and finally the SRND Model, with a 76.86\% F1-score. Thus, resulting in a 81.44\% F1-score for the system, as we can see on Table \ref{table:basleine_results}.

\subsection{Comparison with previous results}
Regarding previous work done on legal text by Humphreys et al.~\citeyearpar{humphreys2020populating}, even though our system results and theirs are not directly comparable (different corpus, different number of samples and some different labels), there are some common labels whose results can be compared. The following tables contain the F1-scores for the concepts extracted by our model and by Humphreys et al.~\citeyearpar{humphreys2020populating}.
\begin{table}[h!]
\centering
\caption{F1-scores of the norms types}
\begin{tabular}{|c c c|} 
 \hline
 Norm Type & Our model & Humphreys et al model \\ [0.5ex]
 \hline
 Obligation & 88.37\% & 76.33\% \\
 Right & 74.18\% & 100\% \\
 Definition & 68.97\% & 93.10\% \\
 Legal Effect & 100\% & 34.78\% \\
 \hline
\end{tabular}
\label{table:humph_ours}
\end{table}

\begin{table}[h!]
\centering
\caption{F1-scores of the norms elements}
\begin{tabular}{|c c c|} 
 \hline
 Norm Element & Our model & Humphreys et al model \\ [0.5ex]
 \hline
 Scope & 72.96\% & 75.55\% \\
 Definiendum & 77.27\% & 54.55\% \\
 Definiens & 64.71\% & 71.43\% \\
 Action & 69.35\% & 30.36\% \\
 Condition & 81.45\% & 32.09\% \\ 
 Includes & 83.24\% & 90.91\% \\
 Exception & 80\% & 22.22\% \\
 \hline
\end{tabular}
\label{table:humph_ours2}
\end{table}

Their system achieved 81.60 F1-score in detecting the norm type, while our norms model achieved a 84.74\%. As we can see from looking at tables \ref{table:humph_ours} and \ref{table:humph_ours2}, their worst norm was ``Legal Effect" with 34.78\% F1-score, while our worst was \textsc{def} with 68.97\%. Regarding the elements of some of their norm types, for those that are similar to ours (v.g. ``Scope", \textsc{definiendum}, \textsc{definiens}, ``Includes", ``Action", ``Condition" and \textsc{exception}), only ``Includes" and \textsc{definiens} had a higher F1-score (90.91\% and 71.43\%, respectively) than our corresponding concepts, \textsc{def-inclusion} and \textsc{definiens}. Their worst result was 22.22\% F1-score for the concept \textsc{exception}, which for our \textsc{exception} concept, we achieved a 80\% F1-score.

In conclusion, taking into account the above observations, it does seem that our model achieves better results in general than the one presented in Humphreys et al.~\citeyearpar{humphreys2020populating}.

\subsection{Existing Noise in the Dataset}
\label{noisesection}

As we have mentioned earlier we found noisy labels during the system implementation, which we then corrected. Still, we did not review the whole dataset since that would take a lot of time. After having finished developing our system, we decided to check if there was still noise in the dataset. We decided to collect randomly, for each label, 30 segments and ask the main annotator to verify how many errors were present. This was done, in order to estimate the presence of noise concerning a specific label, so that when evaluating our system we have that information in mind.


We started by doing this sampling for the Norms labels, whose results can be seen in Figure \ref{table:norms_error}. For this group of labels, since a segment always has some norm, and it is relatively simple to see the norm that should be used instead, when found a noisy label, the main annotator, when doing this verification, not only marked the noisy labels but also point out the real label that should have been used instead.


\begin{table}[h!]
\centering
\caption{Errors found for Norms Labels}
\begin{tabular}{|c c c c c c c c|} 
 \hline
 \multicolumn{8}{|c|}{Correct Label} \\
 \multirow{6}{*}{Annotation} & & OBLIG & DEF & RIGHT & INTRO & LEFFECT & ERROR \\
 & OBLIG & 22 & 2 & 2 & 0 & 0 & 13\% \\
 & DEF & 0 & 30 & 0 & 0 & 0 & 0\% \\
 & RIGHT & 0 & 0 & 30 & 0 & 0 & 0\% \\
 & INTRO & 2 & 0 & 0 & 28 & 0 & 7\% \\
 & LEFFECT & 0 & 0 & 0 & 0 & 30 & 0\% \\ 
 \hline
\end{tabular}
\label{table:norms_error}
\end{table}

If we compare the errors present in the norms and the predictions the Norms Model made, we can find many relationships. For example, our Norms model predicted one \textsc{intro} as a \textsc{oblig}. In our sample of 30 segments, for the label \textsc{intro}, we found an error of 7\% related with the label \textsc{oblig}. This could mean that the incorrect predictions the Norms Model made came from noise in the dataset. For the label \textsc{oblig} we found an error of 13\%, from which half comes from the label ``DEF". This also reflects the incorrect predictions the model made, when from 48 spans that corresponded to ``DEFs", 10 were predicted to be ``OBLIGs" (about 2\%).

Regarding the named entities, \textsc{time\_duration} and \textsc{time\_date\_rel\_text} which were the ones with worse results, were also the two named entities for which we found a higher percentage of error. 

\begin{table}[h!]
\centering
\caption{Errors found for NE and SR Labels}
\begin{tabular}{|c c c|} 
 \hline
 \multirow{7}{*}{NE} & Label & Error \\
 \hline
 & LREF & 3\% \\
 & TREF & 0\% \\
 & NE\_ADM & 8\% \\
 & TIME\_DURATION & 12\% \\
 & TIME\_DATE\_REL\_TEXT & 9\% \\ 
 \hline
 \multirow{6}{*}{SR} 
 & DEF-INCLUSION & 2\% \\
 & ACTION &11\% \\
 & CONDITION & 3\% \\
 & CONCESSION & 3\% \\
 & THEME & 16\% \\ 
 \hline
\end{tabular}
\label{table:others_error}
\end{table}



Finally, for the semantic roles, we did not find error associated with the labels \textsc{definiens} and \textsc{purpose}, still that does not mean it does not exist with certainty. We did find error in other labels for which our model performed correctly, which could mean that if there was not any noise the SRND Model could perform even better regarding those labels, or that it simply learned to predict the wrong information as well. 
Thus, the correction of the annotation process is highly important and action to reduce noise must be actively pursued, specially in such a difficult type of annotation task as this one.

\section{Application To Information Retrieval}

At the beginning of this document, we hypothesized that the extraction of relevant concepts from legislative text could help downstream tasks. For this reason, we decided to see if our system improved the performance of the retrieval information system that was being implemented for the legal texts of the Portuguese Consumer Legislation. The system works by returning the 100 legal acts that are deemed relevant for a given query, and is evaluated by calculating the accuracy of the top x results. This accuracy is measured against a ``golden" result provided by Law experts for that query.

We used our Partial Norm Dependency SNR system predictions for a new set of legal texts, to generate a set of questions and answers(QAs). The process of generating these QAs, consisted of the creation of rules having in mind the entities and their relationships. For example, if a segment has an \textsc{oblig}, with an \textsc{action} and a \textsc{experiencer}, it will generate the following QAs:

Q: What should [\textsc{experiencer}] do?
A: [\textsc{action}]

Q: Who  should [\textsc{action}]?
A: [\textsc{experiencer}]

From these QAs, we used the segment-answer pairs (we created question and answer pairs, so that we could use them for a QA system as well) to fine tune the information retrieval system. This way we could compare the results when no fine-tuning was done, or when fine-tuning was done by performing the Inverse Close Task (ICT) proposed by \cite{Taylor1953ClozePA} versus fine-tuning with our generated segment-answer pairs. ICT basically consists of the division of a segment into parts, and the creation of pairs between each part and the segment itself, to represent question and answer pairs. The results are shown in the following table.


\begin{table}[h!]
\centering
\caption{Accuracy Results for the information retrieval system.}
\begin{tabular}{|c c c c|} 
 \hline
 Accuracy & No fine-tuning & Fine-tuning(ICT) & Fine-tuning(Segment-Answer Pairs) \\ [0.5ex] 
 \hline
 TOP 1 & 44\% & 45\% & 51\% \\
 TOP 3 & 74\% & 76\% & 77\% \\
 TOP 5 & 88\% & 74\% & 90\% \\ 
 TOP 12 & 95\% & 95\% & 95\% \\ 
 \hline
\end{tabular}
\label{table:info_retrieval_comb}
\end{table}


Now if we compare the results, we can clearly see that not only did the retrieval system's performance, when fine-tuned using our predictions, outperform the system with no fine-tuning, but also it outperformed fine-tuning using the ICT method.  Thus, these results show that fine-tuning with generated segment-answer pairs, which contain the relevant semantic information our SNR system extracted, does in deed improve the overall performance of an information retrieval system.

\section{Conclusion}
\label{chap:conclusion}

With this paper, we presented an automatic semantic information extraction system responsible for capturing a defined group of relevant semantic concepts (norm types, named entities and semantic roles) present in Portuguese Consumer Legislation. Our system is composed of three models, whose architecture is inspired on the model of Yu, Bohnet and Poesio \citeyearpar{yu2020named}. We implemented and evaluated three different approaches, and showed that having the system predict the semantic roles knowing the norm type information (Partial  Norm Dependency approach) achieves the best results. We also showed that all models of the Partial  Norm Dependency SNR system, achieved a good performance, attaining a 84.74\% F1-score for the Norms Model, 88.67\% for the Named Entity Model, and 76.86\% for the Semantic Role with Norms Model. Based on all the results, and the existent dataset noise we found, we concluded the Partial  Norm Dependency SNR system had a good performance, resulting in a 81.44\% F1-score.
We also showed how using the predictions of our SNR system, we were able to improve an existing information retrieval system by training the model with our predicted knowledge. 

This is the first system of this kind for Portuguese legal text. We showed how the Partial  Norm Dependency SNR system was implemented and how it is able to capture the relevant semantic concepts, allowing legislation to have a more informative representation. With the presented system, any article in the legislation can be represented not only by its text, but also by the main concepts it includes.

In short, we were able to make several contributions, the major ones being: 

(i) The creation of a dataset with the portuguese consumer law annotated with the corresponding norms and ocncepts;

(ii) The creation of our automatic semantic information extraction system, the SNR system;

(iii) And finally, validating the improvement of existing information retrieval systems, by using the information our own system predicted to train those systems;


In spite of our achievements, there is still room for improvement, which future research could address: 
(i) reviewing the dataset and current annotations, in order to reduce the existing noise;
(ii) training and validating the Full Norm Dependency and Partial  Norm Dependency approaches using the predicted labels instead of the gold labels, in order to see if by doing this the second-level models (v.g. the SRND Model and the NEND Model) will be more robust to errors regarding the norm types labels;
(iii) improving the $ATA$ (Average Token Agreement) metric in order to make sure that it is entirely adequate not only at token level, but also at span level, that is, providing the agreement for a corresponding pair of predicted and gold spans.
By making that improvement, the metric could be used to validate the training of the models and to compare different the approaches, instead of using the F1-score alone; 
(iv) Finally, we used our predictions to generate simple QAs to improve an existing information retrieval system. Future work, not directly regarding our system but rather its application, could include generating more complex rules to create more informative QA. This could improvement the performance of the information retrieval system.

\bmhead{Acknowledgments}
This work was partially supported by project SLICE PTDC/CCI-COM/30787/2017, by funds received from INCM (Imprensa Nacional-Casa da Moeda) under the context of research project "Descodificar a Legislação", and by FCT - Foundation for Science and Technology through projects UIDB/04326/2020 and  UIDB/50021/2020.






\bibliographystyle{apalike}
\bibliography{sn-bibliography}


\end{document}